%% file: template.tex
\documentclass[a4paper]{article}

\usepackage{arxiv}

\usepackage[utf8]{inputenc} 
\usepackage[T1]{fontenc}    
\usepackage{hyperref}       
\usepackage{url}            
\usepackage{booktabs}       
\usepackage{amsfonts}       
\usepackage{nicefrac}       
\usepackage{microtype}      
\usepackage{lipsum}		
\usepackage{graphicx}
\usepackage{natbib}
\usepackage{doi}
\usepackage{latexsym}
\usepackage{amsmath}
\usepackage{amssymb}
\usepackage[mathscr]{eucal}
\usepackage{graphicx,color}
\usepackage{mathtools}
\usepackage{stmaryrd}
\usepackage{url}
\usepackage{xypic}

\usepackage{bm}

\input{notation_defs}

\usepackage{bbm}
\usepackage{tikz-cd}
\usepackage{psfrag}

\def \RR {{\mathbb R}}

\usepackage{amsthm}

\newtheorem{assumption}{Assumption}

\newtheorem{remark}{Remark}

\title{
A Simple Algebraic Solution for Estimating the Pose of a Camera from Planar Point Features}

    \author{ \href{https://orcid.org/0000-0003-1116-7415}{\includegraphics[scale=0.06]{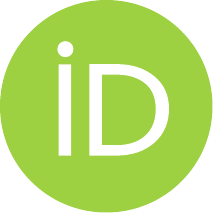}\hspace{1mm}Tarek Bouazza} \\
	I3S, CNRS, Université Côte d'Azur\\
    Sophia Antipolis, France \\
	\texttt{bouazza@i3s.unice.fr} \\
	\And
	\href{https://orcid.org/0000-0002-7779-1264}{\includegraphics[scale=0.06]{orcid.pdf}\hspace{1mm}Tarek Hamel} \\
    I3S, CNRS, Université Côte d'Azur\\
    and Insitut Universitaire de France \\
    Sophia Antipolis, France \\
	\texttt{thamel@i3s.unice.fr} \\
    \And
     \href{https://orcid.org/0000-0002-3769-6174}{\includegraphics[scale=0.06]{orcid.pdf}\hspace{1mm}Claude Samson} \\
	INRIA Sophia Antipolis and I3S\\
    Sophia Antipolis, France \\
	\texttt{csamson@i3s.unice.fr} \\
}

\renewcommand{\headeright}{Author accepted version}
\renewcommand{\undertitle}{Author accepted version\thanks{Bouazza, T., et al. (2025),``A Simple Algebraic Solution for Estimating the Pose of a Camera from Planar Point Features", \textit{Accepted for presentation at IEEE/RSJ International Conference on Intelligent Robots and Systems (IROS)}.

\noindent
\copyright2025 IEEE. Personal use of this material is permitted. Permission from IEEE must be obtained for all other uses, in any current or future media, including reprinting/republishing this material for advertising or promotional purposes, creating new collective works, for resale or redistribution to servers or lists, or reuse of any copyrighted component of this work in other works.
}}

\renewcommand{\shorttitle}{}

\hypersetup{
pdftitle={A template for the arxiv style},
pdfsubject={q-bio.NC, q-bio.QM},
pdfauthor={David S.~Hippocampus, Elias D.~Striatum},
pdfkeywords={First keyword, Second keyword, More},
}

\begin{document}
\maketitle

\begin{abstract}
This paper presents a simple algebraic method to estimate the pose of a camera relative to a planar target from $n \geq 4$ reference points with known coordinates in the target frame and their corresponding bearing measurements in the camera frame. 
The proposed approach follows a hierarchical structure; first, the unit vector normal to the target plane is determined, followed by the camera's position vector, its distance to the target plane, and finally, the full orientation. To improve the method's robustness to measurement noise, an averaging methodology is introduced to refine the estimation of the target's normal direction. 
The accuracy and robustness of the approach are validated through extensive experiments.
\end{abstract}

\section{Introduction}
	
Vision-based motion estimation is a central issue in robotics that involves reconstructing the pose, i.e., position and orientation, of a mobile camera.
Given correspondences between the 3D coordinates of $n$ reference points in an object frame and their image projections, this problem is known as the Perspective-$n$-Point (P$n$P) problem, as introduced by Fischler and Bolles \cite{fischler1981random}. Solutions to the P$n$P problem are typically categorized according to the number of reference points. 
The simplest case, the P3P problem ($n=3$) has been extensively studied \cite{fischler1981random,gao2003complete,kneip2011novel}.
At best (when the points are non-collinear), there are four possible solutions \cite{zhang2005general}. A fourth non-coplanar point is required to resolve the ambiguity and uniquely determine the pose. 
 
For $n \geq 4$, the problem usually admits a unique solution, except in some degenerate configurations (e.g., collinear or coplanar points).
Methods which solve the general P$n$P problem exploit the redundancy of more correspondences to achieve higher accuracy. 
Early non-iterative approaches provide explicit closed-form solutions to compute the camera pose directly \cite{quan1999linear}. 
Iterative algorithms have also been developed to determine the camera pose by minimizing an appropriate cost function (e.g., the reprojection error of the points), such as 
\cite{haralick1989pose}, the POSIT iterative algorithm \cite{dementhon1995model}, the RPP introduced by Lu \emph{et al.} \cite{lu2000fast}, and the SDP method \cite{schweighofer2008globally}.
The drawback of these latter methods is their high
computational cost and the need for a good initialization to avoid local minima.
The EP$n$P method proposed by Lepetit {\em et al.} \cite{lepetit2009ep}
was among the earliest efficient non-iterative solutions that achieve a computational cost of $O(n)$ by expressing the $n$ reference points as a weighted sum of four virtual control points. Subsequent methods with linear complexity, such as DLS \cite{hesch2011direct}, RP$n$P \cite{li2012robust}, and UP$n$P \cite{kneip2014upnp},
were developed to improve the precision by replacing the linear formulation with polynomial solvers. 
The nonstationary case of the P$n$P problem was investigated by the co-authors in \cite{hamel2017riccati}, where the point correspondences were combined with the camera's rigid motion and its velocity measurements to design continuous-time filters.

The P$n$P problem is typically classified into planar and nonplanar cases. 
Apart from recent non-iterative methods like EP$n$P that include special cases when the $n$ points are coplanar, most P$n$P algorithms are not suitable for planar configurations and tend to perform suboptimally. 
Solutions that were specifically designed for the planar P$n$P problem include the extension of the POSIT algorithm \cite{dementhon1995model} to coplanar points \cite{oberkampf1996iterative} and the solution proposed in \cite{schweighofer2006robust} which extends the RPP method \cite{lu2000fast}.
Alternatively, solvers that assume coplanarity, such as the IPPE \cite{collins2014infinitesimal}, are based on the plane-to-image homography transformation \cite{hartley2003multiple} and take advantage of the planar structure to achieve higher accuracy.
Another common strategy to plane-based pose estimation is homography decomposition, where the homography matrix is first computed, e.g. using the Direct Linear Transform (DLT) method \cite{hartley2003multiple}, and then factorized to recover the camera pose using either SVD-based \cite{faugeras1988motion,zhang19963d} or analytical methods \cite{malis2007deeper}.

This paper presents a simple algebraic method to compute the pose of a camera relative to a planar target, as well as determining the distance and normal vector of the target plane, using the coordinates of $n \geq 4$ reference points in the target frame and their corresponding bearing measurements in the camera frame. 
The method follows a hierarchical structure in retrieving the relative pose parameters, the unit vector normal to the target plane is first estimated from one of several sets of four points, then the camera’s position vector and distance to the plane are recovered, and finally its orientation.
A key advantage of the proposed formulation is its ability to recover partial pose information (e.g., direction of the position vector) even when the camera is considerably far from the target and the normal, and subsequently the orientation, are poorly estimated. This direction information could be exploited in control applications, such as the approach to and landing on a planar target, by providing an initial estimate to align the camera with the target frame.
Additionally, when combined with an onboard IMU, this method can be extended to design a filter for the estimation of the inertial velocity relative to a slowly accelerating target. 

The remainder of the paper is organized as follows. Section \ref{sec:prelims} introduces the preliminary notation and Section \ref{sec:algorithm} formally states the problem and outlines the key stages of the proposed algorithm. 
Section \ref{sec:experiments} presents the experimental validation, followed by concluding remarks in Section \ref{sec:conclusion}.

\section{Preliminary notation} \label{sec:prelims}
Throughout the paper, ${\bm E}^3$ denotes the 3D Euclidean vector space and vectors in ${\bm E}^3$ are denoted with bold letters. The associated reference frame is an inertial frame with respect to (w.r.t.) which all other frames are defined.
The inner product in ${\bm E}^3$ is denoted by ``$\cdot$''.
The following notation is used.
\begin{itemize}
\item For $x \in \RR^n$ (resp. $\bm x \in {\bm E}^3$) $|x|$ (resp. $|\bm x|$) denotes the Euclidean norm, and $x^{\top}$ denotes the transpose of $x$.
\item $C$ denotes the camera's optical center;
\item ${\mathcal F}_c=\{C;\bm \imath_c, \bm \jmath_c, \bm k_c \}$ is the camera frame with $\bm k_c$ orthogonal to the camera's image plane;
\item ${\mathcal F}_t=\{O;\bm \imath_t, \bm \jmath_t, \bm k_t\}$ is a frame attached to the target, with the origin $O$ on the target plane and $\bm k_t$ orthogonal to the target plane;
\item $\mathbf{SO}(3):= \{R \in \RR^{3 \times 3} \,|\, R^\top R = I_3, \mathrm{det}(R) = 1\}$ is the special orthogonal group of 3D rotation.
\item $R \in \mathbf{SO}(3)$ is the rotation matrix from ${\mathcal F}_t$ to ${\mathcal F}_c$, i.e. $(\bm \imath_c, \bm \jmath_c, \bm k_c)=(\bm \imath_t, \bm \jmath_t, \bm k_t)R$. 
\item $\xi \in \RR^3$ is the 3d-vector of coordinates of the camera position expressed in the basis of the camera frame, i.e. $\vec{OC}= (\bm \imath_c, \bm \jmath_c, \bm k_c)\xi$. 
\item $x_i \in \RR^2$ is the 2d-vector of coordinates of the point $P_i$ on the target plane, expressed in the basis of the target frame.
The corresponding coordinates in $\RR^3$ are given by
$\mathring{x}_i = (x_i^{\top},0)^{\top}$, i.e. $\vec{OP_i}= (\bm \imath_t, \bm \jmath_t, \bm k_t) \mathring{x}_i$. The homogeneous coordinates of $x_i$ are $\bar{x}_i = (x_i^\top, 1)^\top$.
\item $\mathrm{S}^2 := \{p \in \RR^{3} \,|\, |p| = 1\}$ denotes the unit sphere that consists of all unit vectors in $\RR^{3}$.
\item $e_3=(0,0,1)^{\top} \in \mathrm{S}^2$.
\item $p_i \in \mathrm{S}^2$ is the 3d-unit vector of coordinates of the bearing of the point $P_i$ expressed in the basis of the camera frame, i.e. $\vec{CP_i}/|\vec{CP_i}|=(\bm \imath_c, \bm \jmath_c, \bm k_c)p_i$, computed from the calibrated image coordinates $y_i=(u_x, u_y)$ of $P_i$
as $p_i=  (u_x,u_y,1)^{\top}/\sqrt{u_x^2+u_y^2+1}$. %
\item $\eta \in \mathrm{S}^2$ is the vector of coordinates of the unit vector $\bm k_t$ expressed in the basis of the camera frame. Since $\bm k_t=(\bm \imath_t, \bm \jmath_t, \bm k_t)e_3=(\bm \imath_c, \bm \jmath_c, \bm k_c)R^{\top}e_3$, then $\eta=R^{\top}e_3$.
\item $d$ is the distance between the camera's optical center and the the target plane. Since $d=\vec{CO}\cdot \bm k_t$ with $\vec{CO}=-(\bm \imath_c, \bm \jmath_c, \bm k_c)\xi$ and $\bm k_t=(\bm \imath_c, \bm \jmath_c, \bm k_c)\eta$, then $d=-\eta^{\top}\xi$.
\end{itemize}
\begin{figure}
 \includegraphics[width=.5\textwidth]{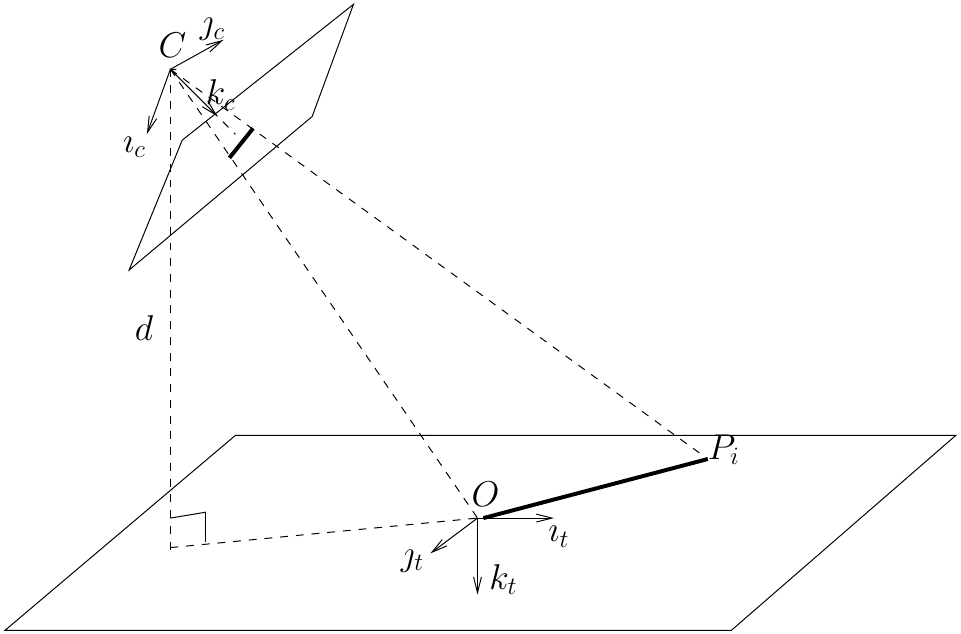}
 \centering
 \caption{Geometric representation of the considered planar P$n$P problem, showing the camera frame $\mathcal{F}_c$, and the planar target frame $\mathcal{F}_t$.}
 \label{figure1}
\end{figure}

\section{Camera pose from matched points}\label{sec:algorithm}
The problem at hand is to recover the camera pose w.r.t. to the target frame, or equivalently $\xi$ and $R$, from a set of known points $P_i$, $i\in \{1,\ldots,n\}$, on the target plane that are seen by the camera and matched with corresponding points on the image plane. We thus assume that the coordinates $x_i$ of the points $P_i$, and the bearings $p_i$ associated with these points, are known (or measured). The problem is thus equivalent to calculating $\xi$ and $R$ from the knowledge of $x_i$ and $p_i$, with $i\in\{1,\ldots,n\}$.
We further assume the following:
\begin{assumption} \label{assumption}
~
\begin{itemize}
\item  $n \geq 4$.
\item There exists a set of four points $P_i$ on the target plane such that no subset of three points is collinear.
\item $\bm k_t \cdot \bm k_c >0$. 
\end{itemize}
\end{assumption}
Note that the third assumption implies that the camera's optical center is located on one side of the target plane, and not on this plane. These assumptions rule out the existence of multiple solutions and of singularities classically associated with the general P$n$P problem. 

The proposed algorithm involves three consecutive stages, namely i) the calculation of $\eta$ from one or several combination of four matched points satisfying the second assumption, ii) the calculation of $d$ and $\xi$, using $\eta$ and four or more points such that three of them are not aligned, and iii) the calculation of $R$, using $\eta$, $\xi$, and four or more matched points such that three of them are not aligned.

From the fact that $\vec{CP_i}=\vec{CO}+\vec{OP_i}$ one easily deduces that $\vec{CP_i}=(\bm \imath_c, \bm \jmath_c, \bm k_c)(R^{\top}\mathring{x}_i-\xi)$. By comparing this relation with the definition of $p_i$ it comes that 
$|\vec{CP_i}|p_i=R^{\top}\mathring{x}_i -\xi$.
Moreover, from the definitions of $\eta$, $d$ and $p_i$, we have $\bm k_t \cdot \vec{CP_i}=d$ and $\bm k_t \cdot \vec{CP_i}=|\vec{CP_i}|\eta^{\top}p_i$. Therefore
\[
|\vec{CP_i}|= \frac{d}{\eta^{\top}p_i}.
\]
From previous relations one gets
\begin{equation} \label{equation1}
\frac{d}{\eta^{\top}p_i}p_i=R^{\top}\mathring{x}_i - \xi .
\end{equation}
This is the central equation, on which the proposed algorithm is based, that relates all known quantities ($\mathring{x}_i$, $p_i$) to those that we want to determine 
($\eta$, $d$, $\xi$, $R$).\\

\subsubsection{Calculation of $\eta$}
~\\
Let us consider a set of four points, labeled $P_1$, $P_2$, $P_3$, and $P_4$, such that no subset of three points is collinear on the target plane. Define $\mathring{x}_{ij} := \mathring{x}_j-\mathring{x}_i$, for $i,j\in\{1,\ldots,4\}$. From \eqref{equation1}
\begin{equation} \label{eta1}
d\left(\frac{p_1}{\eta^{\top}p_1}-\frac{p_i}{\eta^{\top}p_i}\right)=R^{\top}\mathring{x}_{1i},~~~i=2,3,4.
\end{equation}
Determine $\lambda_i \in \RR$, $i=2,3,4$, such that $\sum_{i=2}^4 \lambda_i \mathring{x}_{1i}=0$ and $\sum_{i=2}^4 \lambda_i=1$. These coefficients regrouped in the vector $\lambda=(\lambda_2,\lambda_3,\lambda_4)^{\top}$ are, for instance, given by
$\lambda=X^{-1}e_3$,
where $X \in \RR^{3 \times 3}$ is the invertible matrix defined by 
\[
X:=\left[ \begin{array}{ccc} \bar{x}_{12} & \bar{x}_{13} & \bar{x}_{14} 
\end{array} \right].
\]
Using the definition of $\lambda_i$, $i=2,3,4$, one deduces from \eqref{eta1}
\[
d\left(\frac{p_1}{\eta^{\top}p_1}-\sum_{i=2}^4\lambda_i\frac{p_i}{\eta^{\top}p_i}\right)=0
\]
and, since $d\neq0$
\begin{equation} \label{eta2}
p_1=(\eta^{\top}p_1)\sum_{i=2}^4\lambda_i\frac{p_i}{\eta^{\top}p_i},
\end{equation}
with $\eta^{\top}p_1>0$. On the other hand, since the points $P_2$, $P_3$, and $P_4$ are not aligned, the corresponding bearing vectors $p_2$, $p_3$, and $p_4$ are independent in $\RR^3$ and thus constitute a spanning set of $\RR^3$. Therefore, there exists a {\em unique} vector of non-zero coefficients $a=(a_2,a_3,a_4) \in \RR^3$ such that
\begin{equation} \label{eta3}
p_1=\sum_{i=2}^4 a_i p_i.
\end{equation}
Defining the invertible matrix $B:=[p_2~p_3~ p_4]$, the vector $a$ is given by $a=B^{-1}p_1$.
By identifying the coefficients in the right-hand side of \eqref{eta2} and \eqref{eta3} one gets
\[
a_i= (\eta^{\top}p_1)\lambda_i\frac{p_i}{\eta^{\top}p_i},~~~i=2,3,4.
\]
This also implies that
$p_i^{\top}\eta=(\eta^{\top}p_1)\lambda_i/a_i$, $i=2,3,4$,
or, equivalently
\begin{equation} \label{eta4}
B^{\top}\eta=(\eta^{\top}p_1)b,
\end{equation}
with $b=(\lambda_2/a_2,\lambda_3/a_3,\lambda_4/a_4)^{\top}$.
Therefore, given  that $\eta$ is a unit vector and that $\eta^{\top}p_1>0$, one obtains
\begin{equation} \label{eta5}
\eta=B^{-\top}b/|B^{-\top}b|.
\end{equation}

Note that any other set of four points (with no three collinear points) can be used to compute $\eta$. 
If the bearings $p_i$ were exact projections of the matched image points $y_i$, all estimates of $\eta$ would coincide.
However, in practice, these measurements are imperfect because of pixel noise. For this reason, it may be useful to collect as many estimates $\hat{\eta}_j$ as there are different sets of four matched points and calculate a final estimate of $\eta$ as a weighted sum of $\hat{\eta}_j$, i.e. 
\begin{equation} \label{eq:eta_avg}
\hat{\eta}=\sum_{j=1}^m \gamma_j \hat{\eta}_j/|\sum_{j=1}^m \gamma_j \hat{\eta}_j|,\end{equation}
where $\gamma_j>0$ represents the  weight (confidence) assigned to the $j$th estimate $\hat{\eta}_j$. A natural choice for $\gamma_j$ is given by $\gamma_j = |\min_{i=2,3,4}(a_{i,j})\det(B_j)|$, which gives higher confidence to estimates derived from sets of points that are more widely distributed in the planar target.

Another option is pointed out next. Let $m$ denote the number of four points sets used to estimate $\eta$, and $j=1,\ldots,m$ the ordering index attributed to these sets. Accordingly, specifying \eqref{eta4} for each set yields 
\[
B_j^{\top}\eta=(\eta^{\top}p_{1,j})b_j~~,~j=1,\ldots,m.
\]
This latter relation in turn implies
\begin{equation} \label{eta6}
D\eta=0,
\end{equation}
with $D \in \RR^{3m \times 3}$ 
defined by
\[
D := \left[ \begin{array}{c} 
\gamma_1 (B_1-p_{1,1}^{\top}b_1) \\ \vdots \\ \gamma_m (B_m-p_{1,m}^{\top}b_m)
\end{array} \right].
\]
Equation \eqref{eta6} indicates that $\eta$ is the unit eigenvector associated with the null eigenvalue of the semi-positive matrix $D^{\top}D \in \RR^{3 \times 3}$. This suggests setting $\hat{\eta}$ as the normalized eigenvector associated with the smallest eigenvalue of $D^{\top}D$, which may be non-zero due to pixel noise.   
\begin{remark} \label{remark:averaging_eta}
Directly computing $\hat{\eta}$ as a weighted sum in $\RR^3$ can lead to errors, especially when individual estimates are affected by measurement noise. 
An averaging methodology that guarantees robustness to noise and smoother convergence (called smooth averaging) accounts for the fact that $\eta$ lies on the unit sphere $\mathrm{S}^2$.
Given estimates $\hat{\eta}_j$, $j=1,\dots,m$, from a new image, the prior $\hat{\eta}$ is refined by
\begin{equation}
     \label{eq:eta_avg2} \hat{\eta} \leftarrow \mathrm{exp}(\sigma^\times) \hat{\eta}, \quad \sigma = - \sum_{j=1}^m \gamma_j (\hat{\eta} \times \hat{\eta}_j), 
\end{equation}
where $\mathrm{exp}( (\cdot)^\times) : \RR^3 \rightarrow \mathbf{SO}(3)$ is the matrix exponential map computed using Rodrigues' formula \cite{hartley2003multiple}, and $\hat{\eta}(0)$ 
can be taken from the first image according to \eqref{eq:eta_avg}.
This update corresponds to the solution of the minimization problem
\[\argmin_{\eta \in \mathbb{S}^2} \frac{1}{2}  \sum_{j=1}^m \gamma_j | \hat{\eta}_j - \eta |^2.\]
Alternatively, the update \eqref{eq:eta_avg2} can be applied sequentially for each measurement $\hat{\eta}_j$ obtained from a set of four points to refine $\hat{\eta}$ as
$ \hat{\eta} \leftarrow \mathrm{exp}(\gamma_j (\hat{\eta} \times \hat{\eta}_j)^\times) \hat{\eta}$,  $j=1,\dots,m$. 
However, this comes at the cost of increased computation, since the exponential map must be evaluated $m$ times rather than once.
\end{remark}

\subsubsection{Calculation of $d$ and $\xi$}
~\\
Assume now that $\eta$, or an accurate estimate of it, is known.
Consider $n \geq 4$ matched points $P_i$, $i=1,\ldots,n$, such that at least three are non-collinear. Determine a set of coefficients $\mu_i$, $i=1,\dots,n$, such that
$\sum_{i=1}^n \mu_i x_i=0$ and $\sum_{i=1}^n \mu_i=1$. Regrouped into the $n$-dimensional vector $\mu=(\mu_1,\ldots,\mu_n)^{\top}$, a (non unique) set of such coefficients is given by
$\mu = \bar{X}^{\dag}e_3$,
where $\bar{X} \in \RR^{3 \times n}$ is defined by
\[
\bar{X}:=\left[ \begin{array}{ccc}
\bar{x}_1 & \cdots & \bar{x}_n 
\end{array} \right],
\]
and $\bar{X}^{\dag}=\bar{X}^{\top}(\bar{X}\bar{X}^{\top})^{-1}$ is a pseudo-inverse of $\bar{X}$. Note that $\bar{X}\bar{X}^{\top} \in \RR^{3 \times 3}$ is invertible due to the assumption on the chosen set of matched points. 
Now, define and compute the following vector
\begin{equation} \label{r}
r:=\sum_{i=1}^n \frac{\mu_i}{\eta^{\top}p_i}p_i .
\end{equation}
One deduces from \eqref{equation1} that
\begin{equation} \label{xi1}
\xi=-d r.
\end{equation}
From \eqref{equation1}, one has also
\[
R^{\top} \mathring{x}_i = \xi+\frac{d}{\eta^{\top}p_i}p_i,~~~\forall i=1,\ldots,n.
\]
Therefore,  
$R^{\top} \mathring{x}_i =d q_i$ 
with
\begin{equation} \label{q}
q_i:=\frac{1}{\eta^{\top}p_i}p_i-r.
\end{equation}
and
\begin{equation} \label{d1}
d=\frac{|\mathring{x}_i|}{|q_i|},~~~\forall i=1,\ldots,n.
\end{equation}
From the previous three relations 
\begin{equation} \label{R0}
R^{\top} \frac{\mathring{x}_i}{|\mathring{x}_i|} = \frac{q_i}{|q_i|}.
\end{equation}
Note that relation \eqref{d1} also implies that
\begin{equation} \label{d2}
d=\frac{1}{n}\sum_{i=1}^n \frac{|\mathring{x}_i|}{|q_i|}.
\end{equation}
Due to imperfect measurements in the image, this averaged solution, or a weighted variant, will be preferred in practice. Once $d$ is determined, $\xi$ can be directly obtained using \eqref{xi1}. 

\begin{remark}
When the distance between the camera and the origin of the target plane becomes large, all matched points bearings $p_i$ tend to $-\xi/|\xi|$. This is coherent with relations \eqref{r} and \eqref{d1} according to which $r=-\xi/|\xi|$. This is true even when $\eta$, and subsequently the orientation of the camera w.r.t. the target frame, are no longer well estimated due to image noise and poor conditioning of the matrix $B$. The fact that $r$ continues to be a good estimate of the direction $-\xi/|\xi|$ is a practical asset of the proposed algorithm that can, for instance, be exploited to control the approach of a camera-equipped-vehicle towards the origin of the target plane when the inertial orientation of the vehicle is independently estimated by using IMU measurements.   
\end{remark}

\subsubsection{Calculation of $R$}
~\\
For this third stage, assume that $\eta$ and $\xi$ are known. 
As before, consider a set of $n\geq 4$ matched points $P_i$, $i=1,\ldots,n$, such that at least three of them are non-collinear. Note that there is no requirement of choosing the same set of points as those used in previous stages. As a matter of fact, any two matched points not collinear with the origin of the target frame are sufficient. Define the matrix 
\begin{equation} \label{Xbarbar}
\bar{\bar{X}}:= \left[ \frac{\mathring{x}_1}{|\mathring{x}_1|} ~~ \cdots ~~ \frac{\mathring{x}_n}{|\mathring{x}_n|} ~~ e_3
\right] \in \RR^{3\times n+1},
\end{equation}
which is of rank three because of the assumption on the chosen set of matched points. From \eqref{R0} and the definition of $\eta$, i.e. $\eta=R^{\top}e_3$, $R^{\top} \bar{\bar{X}}=Y$,
with
\begin{equation} \label{Y}
Y:=\left[ \frac{q_1}{|q_1|} ~~ \cdots ~~ \frac{q_n}{|q_n|} ~~\eta \right].
\end{equation}
Therefore
\begin{equation} \label{R}
R^{\top}=Y\bar{\bar{X}}^{\dag},
\end{equation}
with $\bar{\bar{X}}^{\dag}:=\bar{\bar{X}}^{\top}(\bar{\bar{X}}\bar{\bar{X}}^{\top})^{-1}$. 

Measurement noise can cause the computed matrix $R^{\top}$ to deviate from a true rotation matrix in $\mathbf{SO}(3)$. To correct this, a regularization step is needed.
For instance, a Gram-Schmidt orthonormalization procedure can be applied, using the fact that the last column of $R^{\top}$, i.e. $R^{\top}e_3$, is equal to $\eta$.

\begin{remark}
    Similar to Remark \ref{remark:averaging_eta}, the orientation $R$ can be estimated through a smooth averaging procedure rather than an algebraic computation, exploiting the relation $\eta = R^\top e_3$.
    In this case, the full orientation can be determined from the additional knowledge of the $n$ vectors $q_i/|q_i|$ and relation \eqref{R0}.
    This step is omitted here, as the resulting yaw estimates are already sufficiently accurate for the intended purpose.
\end{remark}

\begin{remark}
The homography matrix, which relates the bearings of matched points between two images of a planar structure \cite{hartley2003multiple}, can also be derived from \eqref{equation1}. 
For the bearings $p_{i,1}$ and $p_{i,2}$ of a point $P_i$ observed in two images, one has
\[
\frac{d_j}{\eta_j^{\top}p_{i,j}}p_{i,j}=R_j^{\top}(\mathring{x}_i - \xi_j^t)~~,~j=1,2
\]
with $\xi_j^t$ denoting the coordinates of $\vec{CP_i}$ expressed in the target frame. From this latter relation
\[
R\frac{d_1}{\eta_1^{\top}p_{i,1}}p_{i,1}-\frac{d_2}{\eta_2^{\top}p_{i,2}}p_{i,2}=-\tilde{\xi}
\]
with $\tilde{R}=R_2^{\top}R_1$, the relative rotation between the two camera frames, and $\tilde{\xi}=R_2^{\top}(\xi_1^t-\xi_2^t)$, 
the relative translation vector expressed in the frame of image 2.
Multiplying both members of this latter equality by $\frac{\eta_1^{\top}p_{i,1}}{d_1}$ yields
\begin{equation} \label{homography}
\gamma_i~p_{i,2}=H~p_{i,1}
\end{equation}
where $H:= \tilde{R} + \frac{\tilde{\xi} \eta_1^{\top}}{d_1} \in \RR^{3\times 3}$ is the homography matrix, and $\gamma_i=\frac{\eta_1^{\top}p_{i,1}}{\eta_2^{\top}p_{i,2}}\frac{d_2}{d_1}$. Relation \eqref{homography} indicates that $H$, which may be assimilated to an element of the special linear group $\mathbf{SL}(3)$ (an $8$-dimensional matrix Lie group) after re-scaling to $\bar{H}:=H~\mathrm{det}(H)^{-\frac{1}{3}}$, can be determined from the knowledge of four, or more, bearing pairs $(p_{i,1},p_{i,2})$ \cite{hartley2003multiple}. 
Once estimated, $H$ can in turn be decomposed \cite{malis2007deeper} to extract the rotation $\tilde{R}$, the unit normal $\eta_1$, and the 
and the scaled displacement $\frac{\tilde{\xi}}{d_1}$.
\end{remark}

\section{Experimental validation} \label{sec:experiments}

\begin{figure}[h]
    \centering
    \includegraphics[scale=.232]{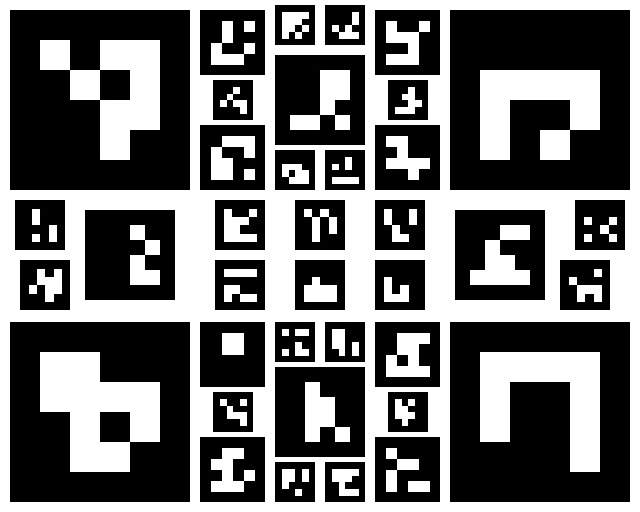} \hspace{.2cm}
    \includegraphics[scale=.05]{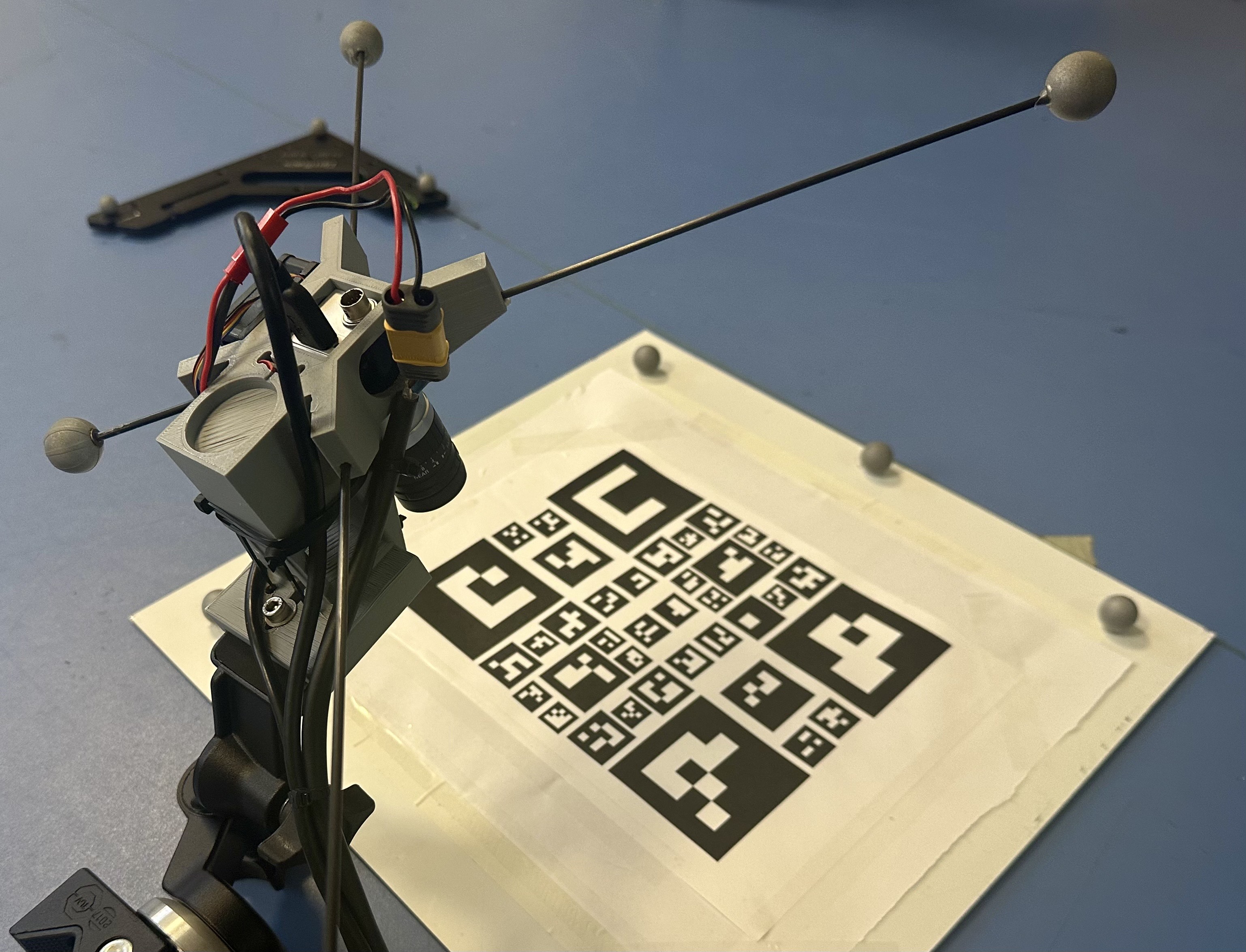} 
    \caption{(left) The arrangement of fiducial markers on the planar target. (right) The camera and target with attached MoCap markers.}
    \label{fig:exp_setup}
\end{figure}

In this section, we present experimental results to demonstrate the performance of the proposed algorithm on real image data. The experiments were performed using a calibrated Basler ace camera that captures images at $20$ frames per second with a resolution of $640 \times 512$ pixels. The point detection was performed with the ArUco library \cite{romero2018speeded} included in OpenCV, which provides reliable corner detection from fiducial markers. The target consisted of a set of 38 fiducial markers (dictionary: DICT\_4X4\_50) of varying sizes arranged on a $ 23.5 \times 18.8$ cm planar board (Fig. \ref{fig:exp_setup}). 

The algorithm was implemented in C++ using the Eigen matrix library. The ground-truth pose measurements were obtained using an OptiTrack motion capture (MoCap) system at a rate of $100$Hz, with markers attached to both the camera and the target (the setup is shown in Fig. \ref{fig:exp_setup}). 
The key advantage of using the ArUco markers is that a single marker provides enough correspondences, via the coordinates of its four corner, to compute the camera's relative pose. 

This setup replicates the scenario of a drone equipped with a camera observing a planar landing marker, where it initially starts from a distant position and then performs a range of motion and orientation changes as it gradually moves closer to the target. 
While the target was downscaled by a factor of four for the experiment due to the MoCap system’s limited tracking volume, in a real deployment, the estimated position would scale with the target's actual size.

\subsection{Results and discussion}
\begin{figure}[!h]
\centering
    \includegraphics[scale=.42]{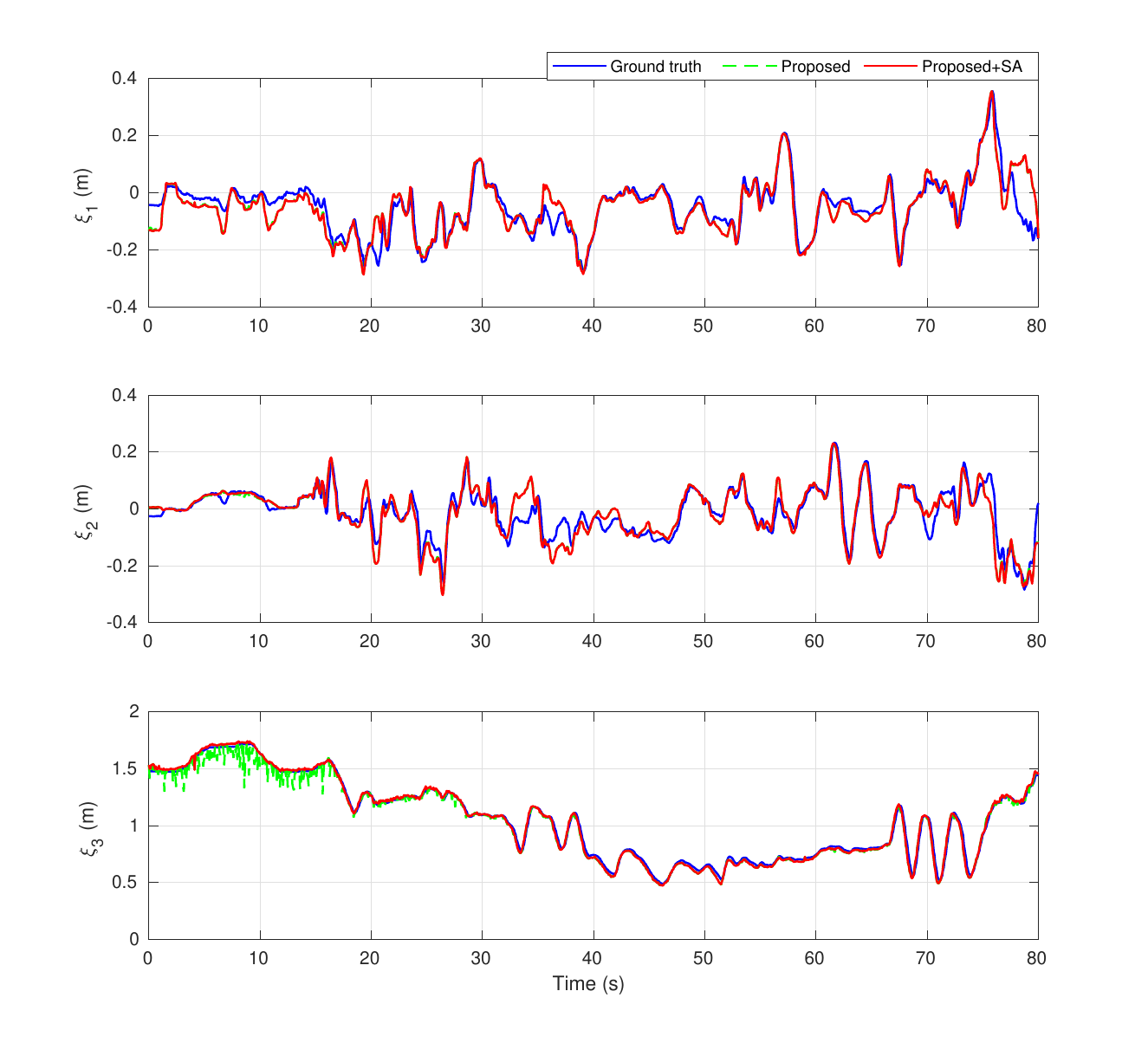} \vspace{-.7cm}
    \caption{Camera relative positions w.r.t the target given by the proposed algorithm without smoothing (green), with the smooth averaging (red),  and MoCap ground truth (blue).}
    \label{fig:estimates_pos_avg}
\end{figure}

\begin{figure}[!h]
\centering
    \includegraphics[scale=.42]{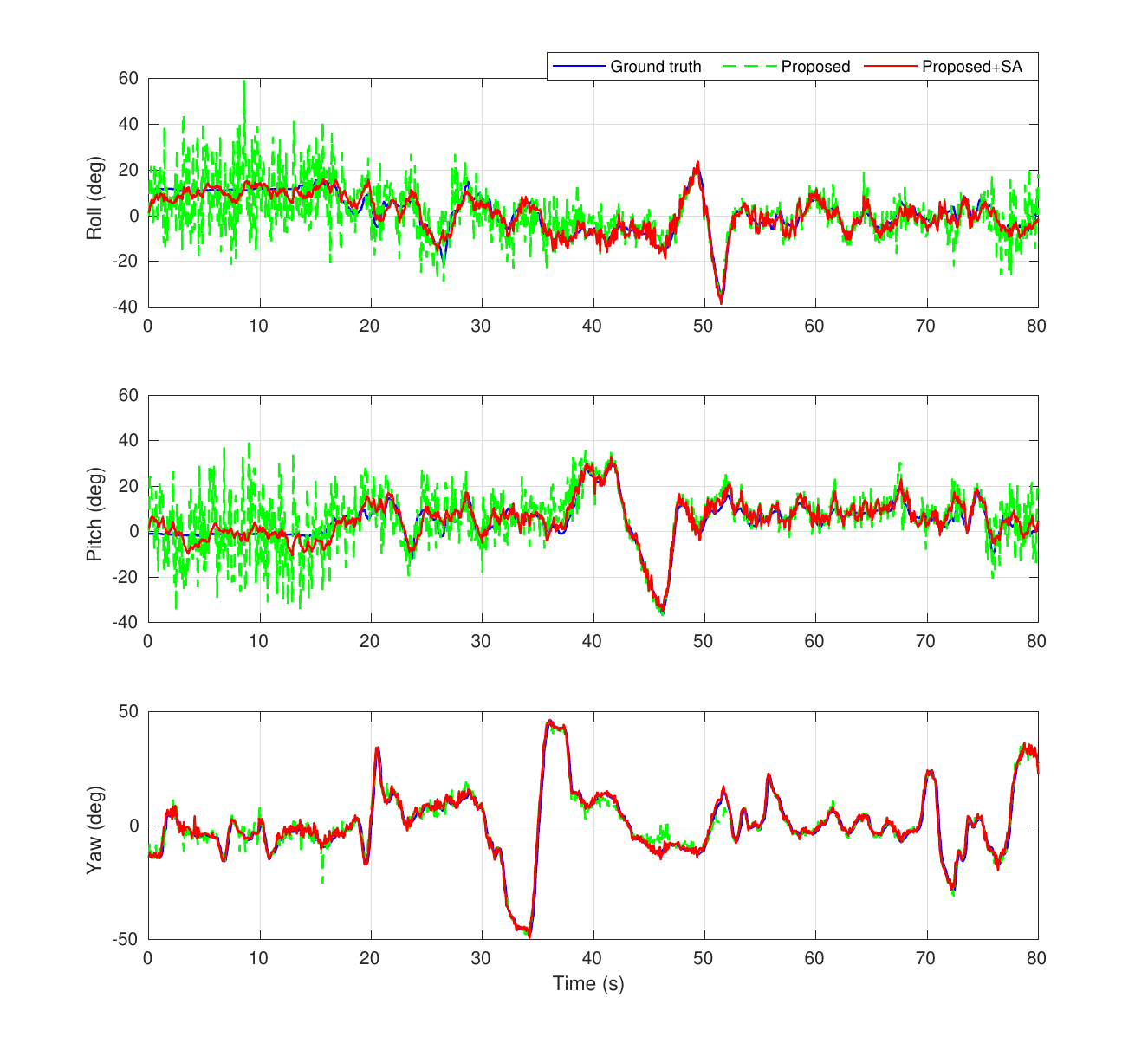} \vspace{-.7cm}
    \caption{Camera relative attitude (Euler angles) w.r.t the target given by the proposed algorithm without smoothing (green), with the smooth averaging (red),  and MoCap ground truth (blue).}
    \label{fig:estimates_orient_avg}
\end{figure}

Figures \ref{fig:estimates_pos_avg} and \ref{fig:estimates_orient_avg} show the estimated camera relative position and attitude (represented by the corresponding Euler angles) obtained using the proposed algorithm with algebraic averaging of $\hat{\eta}$ (relation \eqref{eq:eta_avg}) and the smooth averaging (relation \eqref{eq:eta_avg2}), compared to the ground truth obtained from the MoCap system. 
While the overall motion is consistent with the ground truth, the pitch and roll estimates (related to unit normal vector) provided by the proposed solution with algebraic averaging exhibit significant noise, particularly when the camera is far from the target (i.e., for larger $\xi_3$), as the detected marker corners appear closer in the image. 
Using smooth averaging significantly reduces this noise and improves the attitude estimation.


The estimated pose obtained with the proposed method with smooth averaging was benchmarked against two solvers implemented in OpenCV’s \emph{solvePnP()} function: an iterative P$n$P solver using  the classical Levenberg-Marquardt (LM) optimization method initialized via homography decomposition, and the EP$n$P algorithm.

\begin{figure}[!h]
\centering
    \includegraphics[scale=.4]{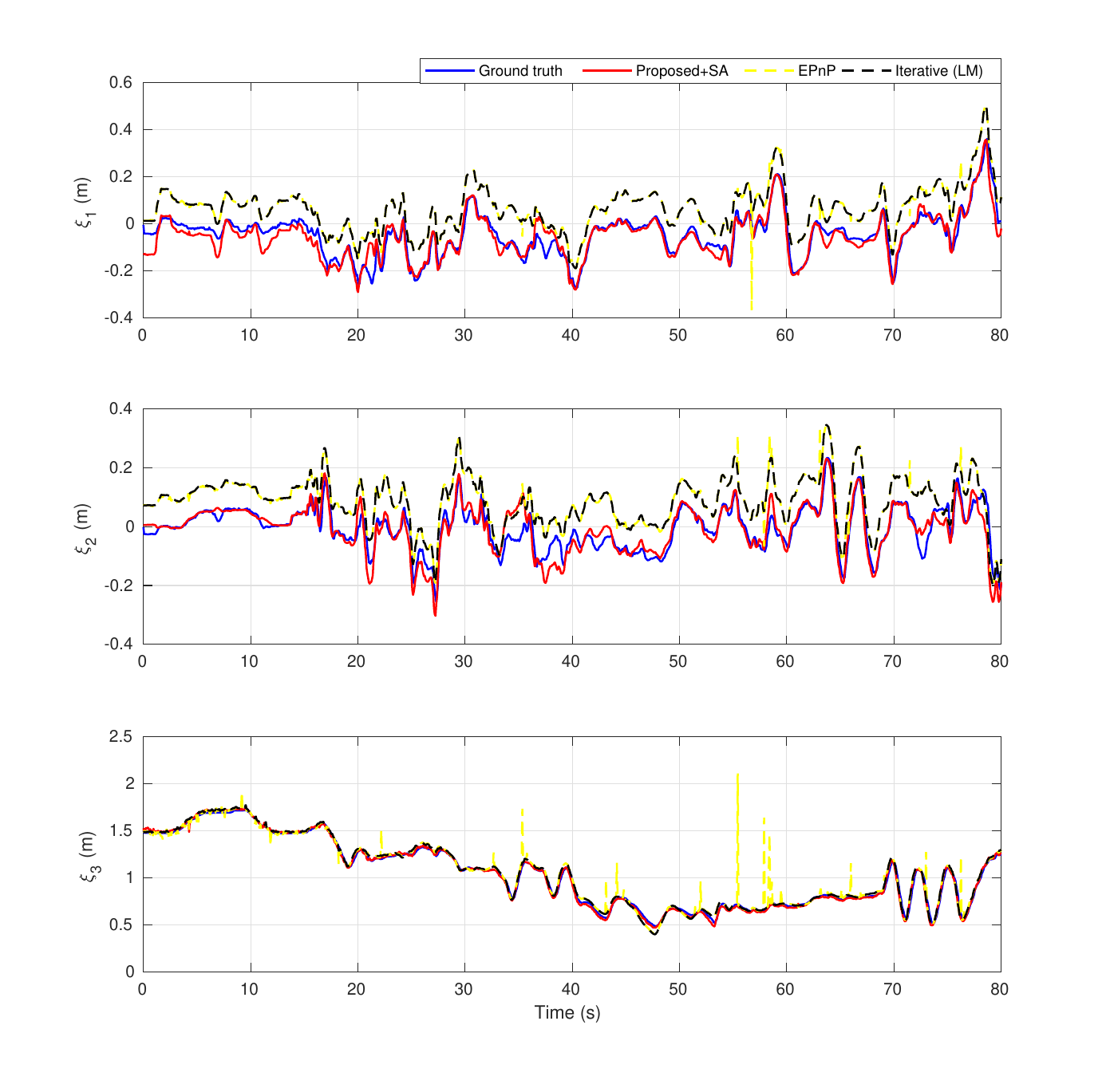} \vspace{-.7cm}
    \caption{Camera relative positions w.r.t the target given by the proposed algorithm (red), the EP$n$P (yellow), iterative P$n$P (black) and MoCap ground truth (blue).}
    \label{fig:estimates_pos_gt}
\end{figure}

\begin{figure}[!h]
\centering
    \includegraphics[scale=.4]{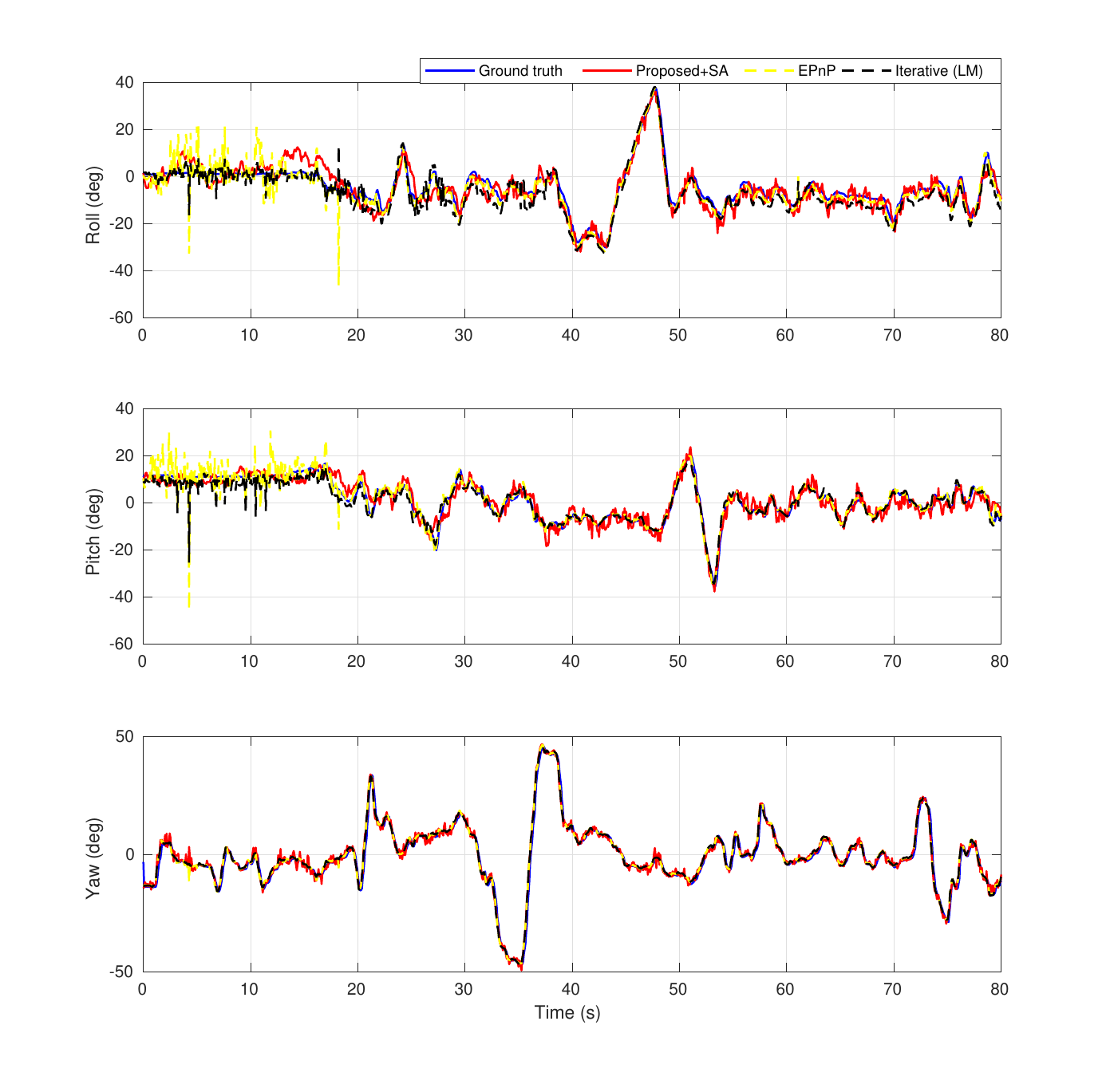} \vspace{-.7cm}
   \caption{Camera relative attitudes (Euler angles) w.r.t the target given by the proposed algorithm (red), the EP$n$P (yellow), iterative P$n$P (black) and MoCap ground truth (blue).}
    \label{fig:estimates_orient_gt}
\end{figure}

Figures \ref{fig:estimates_pos_gt} and \ref{fig:estimates_orient_gt} show the estimated camera position and attitude (Euler angles) obtained using the proposed method with smooth averaging, EP$n$P, and Iterative (LM), compared to the MoCap ground truth.
The results demonstrate that the three methods achieve similar results overall. However, the EP$n$P algorithm occasionally fails to produce valid estimates, and both EP$n$P and LM yield noticeably noisier pitch and roll estimates at larger distances (e.g., $0$-$20$s), the proposed method with smooth averaging provides consistent estimates across the entire sequence.

Table~\ref{tab:pose_rmse} reports the root mean square error (RMSE) for the position and orientation estimates obtained with each method. Orientation errors are computed from Euler angle deviations relative to the MoCap ground truth. Although all methods achieve similar orientation accuracy, with the LM method achieving the lowest angular error, the proposed method yields the lowest position RMSE and significantly outperforms both EP$n$P and LM. 


\begin{table}[h!]
\centering
\caption{Pose estimation RMSE compared to MoCap ground truth. } 
\label{tab:pose_rmse}
\begin{tabular}{lcc}
\toprule
\textbf{Method} & \textbf{Position RMSE} (m) & \textbf{Orientation RMSE} (deg) \\
\midrule
Iterative (LM)     & 0.1574 & \textbf{5.2766} \\
EP$n$P \cite{lepetit2009ep}    & 0.1773 & 5.4194 \\
Proposed & \textbf{0.0623} & 5.3527 \\
\bottomrule
\end{tabular}
\end{table}

\section{Conclusions} \label{sec:conclusion}
In this paper, we addressed the problem of estimating the pose of a camera relative to a planar target from planar point features and their corresponding bearing measurements. The proposed algebraic approach estimates the pose hierarchically and guarantees good estimates of the position direction vector even when the camera orientation is poorly estimated. To improve the method's robustness to pixel noise, a smooth averaging formulation was introduced to refine the estimation of the target's normal. 
Extensive experiments demonstrate the accuracy and reliability of the method. 
The primary motivation for this work is its potential application to vision-based control of autonomous drone landing on planar targets. Future work will focus on extending the approach to moving targets by combining data from an onboard IMU.

    \section*{Acknowledgment}
    This work has been supported by the "Grands Fonds Marins" Project Deep-C, and the ASTRID ANR project ASCAR.


\bibliographystyle{unsrtnat}
\bibliography{ref}

\appendix



\end{document}

%% file: notation_defs.tex
\DeclareMathOperator*{\argmin}{argmin}

\usepackage{accents}
\makeatletter
\providecommand{\scirc}{%
    \hbox{\fontfamily{\rmdefault}\fontsize{0.4\dimexpr(\f@size pt)}{0}\selectfont{\raisebox{-0.52ex}[0ex][-0.52ex]{$\circ$}}}}

\makeatother

\mathchardef\mhyphen="2D

